\documentclass[conference]{IEEEtran}
\IEEEoverridecommandlockouts
\usepackage{cite}
\usepackage{amsmath,amssymb,amsfonts}
\usepackage[linesnumbered, ruled]{algorithm2e}
\SetKwRepeat{Do}{do}{while}
\usepackage{algorithmic}
\usepackage{graphicx}
\usepackage{textcomp}
\usepackage{xcolor}
\def\BibTeX{{\rm B\kern-.05em{\sc i\kern-.025em b}\kern-.08em
    T\kern-.1667em\lower.7ex\hbox{E}\kern-.125emX}}
\begin{document}

\newcommand{\redcolor}[1]{\textcolor{red}{#1}}

\title{Multiple-Pilot Collaboration for Advanced Remote
Intervention using Reinforcement Learning\\

\thanks{This work was supported by the Engineering and Physical Sciences
Research Council (EPSRC) Grant EP/P012779/1; European Commission
Grants H2020 PH-CODING (FETOPEN 829186), CONBOTS (ICT 871803),
REHYB (ICT 871767), NIMA (FETOPEN 899626) and UK EPSRC Grant
FAIR-SPACE EP/R026092/1. (Corresponding author: Weibang Bai)}
}

\author{\IEEEauthorblockN{Ziwei Wang}
\IEEEauthorblockA{\textit{Department of Bioengineering} \\
\textit{Imperial College London}\\
London, United Kingdom \\
ziwei.wang@imperial.ac.uk}
\and
\IEEEauthorblockN{Weibang Bai }
\IEEEauthorblockA{\textit{Department of Computing} \\
\textit{Imperial College London}\\
London, United Kingdom\\
wbbai@imperial.ac.uk }
\and
\IEEEauthorblockN{Zhang Chen}
\IEEEauthorblockA{\textit{Department of Automation} \\
\textit{Tsinghua University}\\
Beijing, China\\
cz\_da@tsinghua.edu.cn }
\and
\IEEEauthorblockN{Bo Xiao}
\IEEEauthorblockA{\textit{Department of Computing} \\
\textit{Imperial College London}\\
London, United Kingdom \\
b.xiao@imperial.ac.uk }
\and
\IEEEauthorblockN{Bin Liang}
\IEEEauthorblockA{\textit{Department of Automation} \\
\textit{Tsinghua University}\\
Beijing, China \\
bliang@tsinghua.edu.cn \vspace{-5mm}}
\hspace{10cm}
\and
\IEEEauthorblockN{Eric M. Yeatman \hspace{3cm}}
\IEEEauthorblockA{\hspace{-1.5cm}\textit{Department of Electrical and Electronic Engineering \hspace{1cm}} \\
\textit{Imperial College London \hspace{3cm}}\\
London, United Kingdom \hspace{3cm}\\
e.yeatman@imperial.ac.uk \vspace{-5mm} \hspace{3cm}}
\hspace{-10cm}
}

\maketitle

\begin{abstract}
The traditional master-slave teleoperation relies on human expertise without correction mechanisms, resulting in excessive physical and mental workloads. To address these issues, a co-pilot-in-the-loop control framework is investigated for cooperative teleoperation. A deep deterministic policy gradient (DDPG) based agent is realised to effectively restore the master operators' intents without prior knowledge on time delay. The proposed framework allows for introducing an operator (i.e., co-pilot) to generate commands at the slave side, whose weights are optimally assigned online through DDPG-based arbitration, thereby  enhancing the command robustness in the case of possible human operational errors. With the help of interval type-2 (IT2) Takagi–Sugeno (T-S) fuzzy identification, force feedback can be reconstructed at the 
master side without a sense of delay, thus ensuring the telepresence performance in the force-sensor-free scenarios. Two experimental applications validate the effectiveness of the proposed framework.
\end{abstract}

\begin{IEEEkeywords}
Collaborative teleoperation, IT2 fuzzy system, Reinforcement learning, Time delay, Kalman filter
\end{IEEEkeywords}

\section{Introduction}
\IEEEPARstart{R}{obotic} assisted remote intervention is widely adopted in extreme manipulating scenarios like outer space, under water, and minimally invasive surgery (MIS), as it extends the operational ability and improve the ergonomics and intelligence through teleoperation schemes \cite{niemeyer1991stable,lawrence1993stability,wbai2021dual,troccaz2019frontiers,Wang2019Adaptive}. 
The human intention and command on the master side can be transferred through communication channels to be reproduced on the slave robots. Meanwhile, it can offer cross-continent intervention or assistance even in the absence of local experienced operators. 

However, time delay induced by long-distance communication will deteriorate the stability for robotic teleoperation  systems\cite{arcara2002control}.
For instance, in the application for robotic assisted MIS\cite{rezazadeh2019robotic,bai2017modular,bai2017novel} and assistive exoskeleton \cite{xue2021Continuous,Xue2019Adaptive}, it has been shown that the time delay should be ideally less than 180ms\cite{Ivanova2021Short} to ensure intuitive haptic communication for human-robot interaction. However, prior knowledge on time delay and its derivative are difficult to obtain in advance. For the communication in cross-regional teleoperation, the time delay can go easily over 180ms \cite{Wang2020Fault-Tolerant,Wang2020Event}. Thus, overcoming large time delay without prior information is still an open challenge for the control design in teleoperation.

To address the resulted control issues, wave-variable-based, passivity-based, predictive-based and adaptive-based control schemes have been reported in the literature\cite{SUN2014Application,nuno2011passivity,uddin2016predictive,Chan2014Application}. But in general, teleoperation relies on the master operator's full control, which is prone to cause fatigue to the human operator and thus may deteriorate the control performance. Moreover, operational errors may randomly happen to the human, which will be directly reproduced at the slave side. In addition, the human operator can only access limited environmental information from the slave side, which can not guarantee the correct intervention all the time.

To overcome these problems, introducing multiple operators with separate interfaces could facilitate the robust arbitration for the master commands. It avoids subjective operation errors effectively and therefore reduce the physical and mental workload for each. Following the multi-operator based concept, the authority weights for different human operators in dual-user or multiple-operators haptic systems can be assigned \cite{shahbazi2018systematic,Khademian2012Dual-User}. However, the authority factors are normally pre-determined as constants making the incorrect commands under large weights cannot be compensated for, weakening the coordination performances of the multiple operators.

In this paper, in order to further improve the overall performance of the multi-operator based approach, we introduce a new deep reinforcement learning (RL) based
multiple-pilot collaboration framework for teleoperation. Besides the normally installed interfaces on the remote master side, we add a co-pilot at the slave side to offer additional intervention for teleoperation. In the proposed control framework, dual Kalman filters (KF) are used to distribute multiple surgeons' commands online. Based on the fused master signal, the deep deterministic policy gradient (DDPG) \cite{lillicrap2015continuous} based RL algorithm is employed to optimize the policy which corrects the phase difference and data loss caused by unknown time delay. Fuzzy identification based force estimation for the robot-environment interaction is developed to predict the force on the salve side. The developed control framework can guarantee the synchronisation tracking performance and the transparency while improving the robustness of decisions. 

\begin{figure}
\centering
\includegraphics[width=0.95\linewidth]{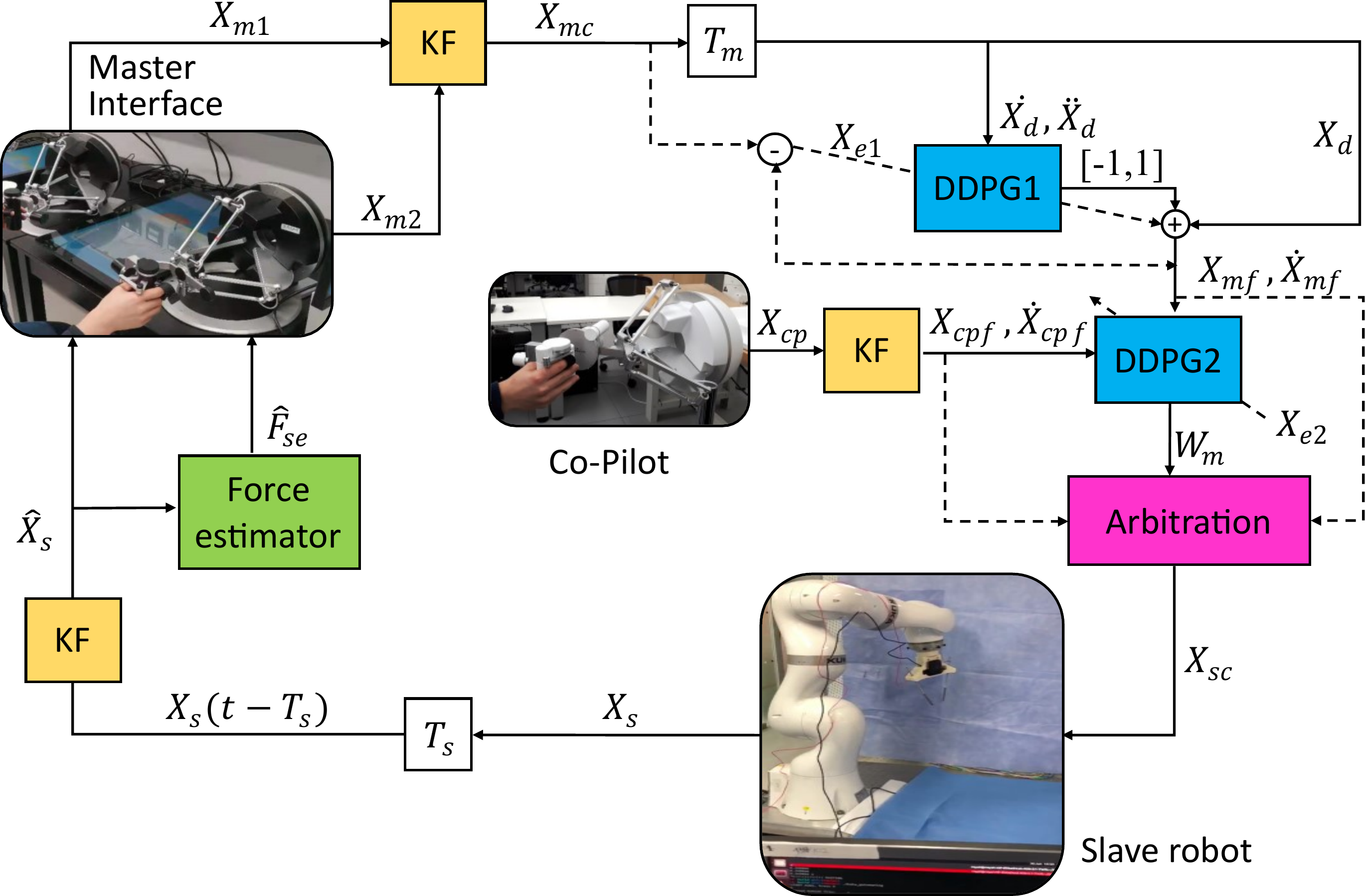}
\vspace{-2mm}
\caption{Multiple-Pilot cooperation framework diagram.}
\vspace{-4mm}
\label{f:diagram}
\end{figure}

\section{Methods}
The overall control framework for multiple-pilot collaboration is illustrated in Fig.\,\ref{f:diagram}, which is composed of dual master operators with visual feedback, one co-pilot, the slave robot and the environment. In what follows, the argument will be omitted for simplicity when not specifically highlighted.

\subsection{Multi-Operator Command Fusion}
Human commands from the dual master interfaces, $X_{m1}$ and $X_{m2}$, are fused through a KF module, whose output is denoted by $X_{mc}$. Consider the process system for KF module used at the master side:
\begin{equation} \label{eq1}
 X_i=AX_{i-1}+\omega_{i}, \ \omega_i \in \mathcal{N}(0,W_i)
\end{equation}
\begin{equation} \label{eq2}
Z_i=CX_i+v_i, \ v_i \in \mathcal{N}(0,R_i)
\end{equation}
where $i\in \mathbb{N}$ is the time-step index with $\mathbb{N}$ being the set of positive integers. $A$ is the system matrix, $X_i$ the system state described in Cartesian space, $\omega_i$ the process noise with covariance $W_i=E[\omega_i\omega_i^T]$, $Z_i$ the observed state, $C$ the projection matrix, and $v_i$ the observation noise with covariance $R_i=E[v_i v_i^T]$.
At each time step, the future value of each state is estimated as in  \eqref{eq3} and then corrected based on the received measurements \eqref{eq4},
\begin{equation} \label{eq3}
 {X}^\ast _i=A\hat{X}_{i-1}
\end{equation}
and
\begin{equation} \label{eq4}
 \hat{X}_i= {X}^\ast _i+K_{i}(Z_i-C  {X}^\ast _i)
\end{equation}
where ${X}^\ast_i$ is the state estimation, $\hat{X}_i$ is the corrected state and $K_i$ is the Kalman gain matrix expressed in \eqref{eq5}-\eqref{eq6},
\begin{equation} \label{eq5}
K_i= P_{i-1}C^T (CP_{i-1}C^T+R_i)^{-1}
\end{equation}
\vspace{-3mm}
  \begin{equation} \label{eq6}
 P_i=A(P_{i-1}-K_iCP_{i-1})A^T+W_i
\end{equation}

\textit{Remark 1}. In addition to the above KF incorporating multiple operator commands, two Kalman filters are employed in the proposed teleoperation framework, satisfying the parameter update process described above. The one near the co-pilot guarantees the smoothness and availability of $X_{cpf}$ and $\dot{X}_{cpf}$. The slave state predicted by the Kalman filter ($\hat{X}_s$) enhances the transparency as it provides a delay-free force feedback through the force estimator.
\subsection{DDPG-based Prediction and Arbitration}
Two DDPG agents are developed to tackle the time delay and decision making issues in the proposed multiple-operator teleoperation. The first agent (DDPG1) is trained to compensate for a specific instantaneous value to the delayed master signal, including the velocity and acceleration ($\dot{X}_{d}$, $\ddot{X}_{d}$) for each Cartesian dimension. Given the instantaneous signal transmission error $X_{e1}=X_{mc}-X_{mf}$, in which $X_{mf}$ is the sum of $X_d$ and the DDPG1 output (also one of the inputs of DDPG2), the optimization of DDPG1 is guided through the properly designed reward function $r_1$, which encourages the networks to minimize $X_{e1}$ caused by forward time delay $T_m$.

DDPG1 is composed of two neural networks (NNs), namely a deep $Q$ network (critic NN) that estimates the state-action values \eqref{value}, and a policy network (actor NN) that generates the optimal action under each state  \eqref{action},
\begin{equation} \label{value}
 \hat{Q}(s_i,a_i)= \sum_{j=1}^{N_Q}\phi_{i}^j\sigma_{cj}(s_i,a_i),
\end{equation}
\vspace{-2mm}
 \begin{equation} \label{action}
 \hat{\mu}(s_i)= \sum_{j=1}^{N_\mu}\theta_{i}^j\sigma_{aj}(s_i),
\end{equation}
where $s_i$ is the state in the NNs, including the velocity and acceleration received at the slave side, namely $s_i=\{\dot{X}_{d_i},\ddot{X}_{d_i}\}$. $\hat{Q}(s_i,a_i)$ is the estimated state-action value for state $s_i$ and action $a_i$ at instant $i$. $\hat{\mu}(s_i)$ is the estimated optimal action for $s_i$, $N_Q$ and $N_\mu$ are the number of basis functions in the output NN layer, $\phi_{i}$ and $\theta_{i}$ are the corresponding network weights for the output layer, and $\sigma_{cj}$ and $\sigma_{aj}$ are the basis functions for the critic and actor networks respectively. A dense model is used as the basis function while the ReLu is chosen as the activation function for both networks. The critic NN minimizes the following cost function 
\begin{equation} \label{cost}
J_Q(\phi)=\frac{1}{N_c}\sum_{i=1}^{N_c}(y_i-Q(s_i,\mu(s_{i})))^T(y_i-Q(s_i,\mu(s_{i}))),
 \end{equation}
\vspace{-2mm}
 \begin{equation} \label{cost2}
y_i=r_i+\gamma Q'(s_{i+1},\mu'(s_{i+1})),
 \end{equation}
where $N_c$ is the number of transitions in the minibatch sampled from the critic NN replay buffer, $y_i$ is the critic network output, $Q$ is the state-action value for the critic NN, $r_i$ is the immediate reward, $\gamma\in(0,1)$ is the discount factor, $Q'$ is the state-action value obtained from the target network, $\mu'$ is the target network policy, and $\mu$ is the actor NN policy. On the other hand, the actor NN updates the network's weights by minimizing the the cost function in \eqref{gradient},
 \begin{equation} \label{gradient}
 J_\mu(\theta)=\frac{1}{N_a}\sum_{i=1}^{N_a}{ Q(s_i,\mu_{s_i}) },
 \end{equation}
where $N_a$ is the number of transitions in the minibatch sampled from the replay buffer. The reward function for DDPG1 is defined as:
 \begin{equation}
     r_{1i}^j = \begin{cases} -10X_{{e1}_i}^j & \text{if} ~X_{{e1}_i}^j>0.1 \\
     2 & \text{if} ~0.01<X_{{e1}_i}^j<0.1\\
     10 & \text{else}\\     \end{cases}
 \end{equation}
where the superscript $j \in \mathbb{N}$ is the $j$-th element index.

The second DDPG agent (DDPG2) aims to allocate weights online for co-pilot and master operators, and outputs the optimised weight matrix ($W_m$) to minimize $X_{e2}$. $X_{e2}$ is the error of the final command ($X_{sc}$) and the task target, which is generated through the arbitration module as following blending form
 \begin{equation}
    X_{sc}=W_mX_{cpf}+(I-W_m)X_{mf},
 \end{equation}
where $W_m$ is a diagonal matrix and each entry is between 0 and 1. $I$ is the identify matrix with appropriate dimension. It also applies to the velocity command in the same manner. Similar with the DDPG1, the reward function with respect to DDPG2 is designed as 
  \begin{equation}
     r_{2i}^j = \begin{cases} -10X_{{e2}_i}^j & \text{if} ~X_{{e2}_i}^j>0.1 \\
     2 & \text{if} ~0.01<X_{{e2}_i}^j<0.1\\
     10 & \text{else}\\     \end{cases}
 \end{equation}

Finally, a state feedback controller such as a PID controller can be used to determine the control command for the slave robot, which updates the end-effector state ($X_{s}$) of the slave robot online.

\textit{Remark 2}. The signals used are all represented in Cartesian space. In this way, the slave robot does not have to comply with the master interfaces in terms of configurations. DDPG1 can ensure the restoration of the master intention without the prior information of time delay, overcoming packet loss and phase delay problems due to the communication channel. In addition, the output of DDPG2 provides the dynamically adjustable weights for arbitration, integrating the intentions of operators at different sides.  

\textit{Remark 3}. The proposed framework allows the 
boundary of weights ($W_m$) to be changed by adjusting the tanh layer parameters of the action NN according to the task requirements. Each entry of $W_m$ should be in the range of $[0,0.5]$ for master-oriented teleoperation tasks. As an example, the driver and the surgeon should play a dominant role in shared driving and telesurgery in terms of human assistance and safety. Regarding the scenarios where the master operator has similar expertise to the co-pilot, the output range of Actor NN can be set as $[0,1]$. The initial weights can be assigned to the co-pilot based on a prior knowledge or confidence level, which are then optimally adjusted by DDPG2 based on his/her performance.

\subsection{IT2 Fuzzy Identification based Force Estimator}
The installation of force sensors may not be impractical in certain scenarios, such as space robot grasping and surgical robot cutting, due to conflicting layout of end-effector tools and force sensors. Therefore, data-based force feedback to the master operator in force-sensor-free scenarios facilitates to improve the transparency of teleoperation. To this end, we develop an IT2 T-S fuzzy identification to 
reconfigure interactive force at the master side.

Since the input data may contain different features, each feature needs to be normalized so that its value domain is in a similar range to facilitate the clustering of interaction environments. In this regard, z-score normalization and linear discriminant analysis (LDA) are exploited for the pre-processing procedure, where LDA is employed to maximise the distance between data in different categories and minimise that in the same category. Let the data preprocessed by the above two steps be $\{\hat{X}_s,F_{se}\}$. In what follows, we adopt the Fuzzy C-Means (FCM) algorithm to identify fuzzy sets as follows
\begin{equation}
\label{eq:FCM1}
\begin{aligned}
&\min_C\quad \sum_{i=1}^N\sum_{j=1}^p {({\omega}_{i}^j)}^m \left\lVert \hat{X}_{si}-c_j\right\rVert^2\\
&\text{subject to}\quad \sum_{j=1}^l {\omega}_{i}^j=1\\
\end{aligned}
\end{equation}
where the subscript $i\in \mathbb{N}$ stands for the $i$-th sampling, $p$ is the number of fuzzy rules extracted by subtractive clustering \cite{chiu1994fuzzy} and $N$ is the number of data points. $m>1$ is fuzzy partition matrix exponent and $ C=\{{c}_{1},...,{c} _{p}\}$. The membership function with respect to Rule $j$ is denoted by ${\omega}_{i}^j$. Taking the Lagrange multiplier $\lambda\in\mathbb{R}_{+}$ into the problem \eqref{eq:FCM1}, one can obtain 
\begin{equation}
    \label{eq:FCM2}
    \mathcal{L}=\sum_{i=1}^N\sum_{j=1}^p ({\omega}_{i}^j)^m \left\lVert \hat{X}_{si}-c_j\right\rVert^2+\lambda\left(1-\sum_{j=1}^p {\omega}_{i}^j\right),
\end{equation}
where the first-order partial derivatives with respect to $\lambda$ and ${\omega}_{i}^j$ are as follows
\begin{equation}
\label{eq:FCM3}
\nabla_{\lambda, {\omega}_{i}^j} \mathcal{L}=0	\Leftrightarrow
\begin{cases} 
m({\omega}_{i}^j)^{m-1}\left\lVert \hat{X}_{si}-c_j\right\rVert^2-\lambda=0\\
1-\sum_{j=1}^p {\omega}_{i}^j=0\\
\end{cases}
\end{equation}

Solving the above equation group yields
\begin{equation}
\label{eq:FCM4}
c_{j}=\frac{\sum_i^N({\omega}_{i}^j)^{m}\hat{X}_{si}}{\sum_i^N {\omega}_{i}^j} 
\end{equation}
and
\begin{equation}
\label{eq:FCM5}
{\omega}_{i}^j=
\begin{cases} 
0& \text{if}\, \hat{X}_{si}=c_k\\
\frac{1}{\sum_{k=1}^p\left(\frac{\left\lVert \hat{X}_{si}-c_j\right\rVert}{\left\lVert \hat{X}_{si}-c_k\right\rVert}\right)^{\frac{2}{m-1}}}& \text{if}\, \hat{X}_{si}\neq c_k\\
1& \text{if}\, \hat{X}_{si}=c_j\\
\end{cases}
\end{equation}

Considering the measurement noise in $\hat{X}_{si}$ and the system uncertainty, the firing strength of the Rule $j$ is within the following interval set \cite{Mendel2006it2}
  \begin{equation}
  \label{eq:IT2Fuzzy1}
  \Omega_i^j(\hat{X}_{si})=[\underline{\omega}_{i}^j(\hat{X}_{si}), \bar{\omega}_{i}^j(\hat{X}_{si})], j\in \mathbb{N}^{+}{[1,p]},
  \end{equation}
  in which $\underline{\omega}_{i}^j(\hat{X}_{si})$ and $\bar{\omega}_{i}^j(\hat{X}_{si})$ denote the lower and upper membership function such that $0\le\underline{\omega}_{i}^j(\hat{X}_{si}) \le \bar{\omega}_{i}^j(\hat{X}_{si}) \le 1$. Hence, the data-driven IT2 T-S fuzzy system is described by
  \begin{equation}
  \begin{aligned}
  \label{eq:IT2Fuzzy2}
  \hat{F}_{sei}=&\sum_{j=1}^p\tilde{\omega}_i^j(\hat{X}_{si})M_i^j\hat{X}_{si}^*\\
  \end{aligned}
  \end{equation}
  where $\hat{X}_{si}^*=[1, \hat{X}_{si}^T]^T$ is the augmented vector of $\hat{X}_{si}^*$. $M_i^j$ is the parameter matrix updated by weighted recursive least squares (WRLS) algorithm, namely 
  \begin{equation}
  \label{eq:WRLS1}
   M_{i+1}^j=M_i^j+K_i\left({F}_{se(i+1)}^T-(\hat{X}_{s(i+1)}^*)^T (M_i^j)^T\right), 
  \end{equation}
  where $K_i= \frac{S_{i+1}\hat{X}_{si}^*}{\sigma+(\hat{X}_{si}^*)^TS_{i}(\hat{X}_{si}^*)}$, $\sigma\in[0,1]$,
  \begin{equation}
  \label{eq:WRLS2}
     S_{i+1}= S_{i}- \frac{S_{i}\hat{X}_{si}^*(\hat{X}_{si}^*)^TS_{i}}{\sigma+(\hat{X}_{si}^*)^TS_{i}(\hat{X}_{si}^*)}
  \end{equation}
  and
  \begin{equation}
  \label{eq:IT2Fuzzy3}
  \tilde{\omega}_j(\hat{X}_{si})=\underline{\omega}_j(\hat{X}_{si})\underline{b}_j(\hat{X}_{si})+\bar{\omega}_{j}(\hat{X}_{si})\bar{b}_j(\hat{X}_{si}),
  \end{equation}
  with $\underline{b}_j(\hat{X}_{si})$, $\bar{b}_j(\hat{X}_{si})\in [0,1]$ such that $\underline{b}_j(\hat{X}_{si})+\bar{b}_j(\hat{X}_{si})=1$.

\begin{algorithm}
  \KwData{Slave states after Kalman filtering on the master side ($\hat{X}_s$)}
  \KwResult{IT2 fuzzy systems and identified clusters}
  Z-score normalize and LDA for the collected $\hat{X}_s$\; 
  Calculate the cluster centers as \eqref{eq:FCM4} and membership functions as \eqref{eq:FCM5}\; 
  \While{$i\le N$}{
    Conduct state augmentation as $\hat{X}_{si}^*$\;
    \Repeat{}{
    Update the parameter matrix as \eqref{eq:WRLS1}-\eqref{eq:WRLS2}\;
    }
    Calculate the IT2 T-S fuzzy output as \eqref{eq:IT2Fuzzy2}\;
  }
  \caption{Data-Driven IT2 T-S Fuzzy Identification Algorithm}
\end{algorithm}

\section{Experiment Results}
We have experimentally validated the individual technologies of the proposed framework and performed post-integration systemic testing, where the setup is composed of two 7-DoF Omega. 7 interfaces (Force Dimension Inc., Switzerland) and an LBR iiwa 14 robot (KUKA Inc., Germany).
\begin{figure}
\centering
\includegraphics[width=8.5cm]{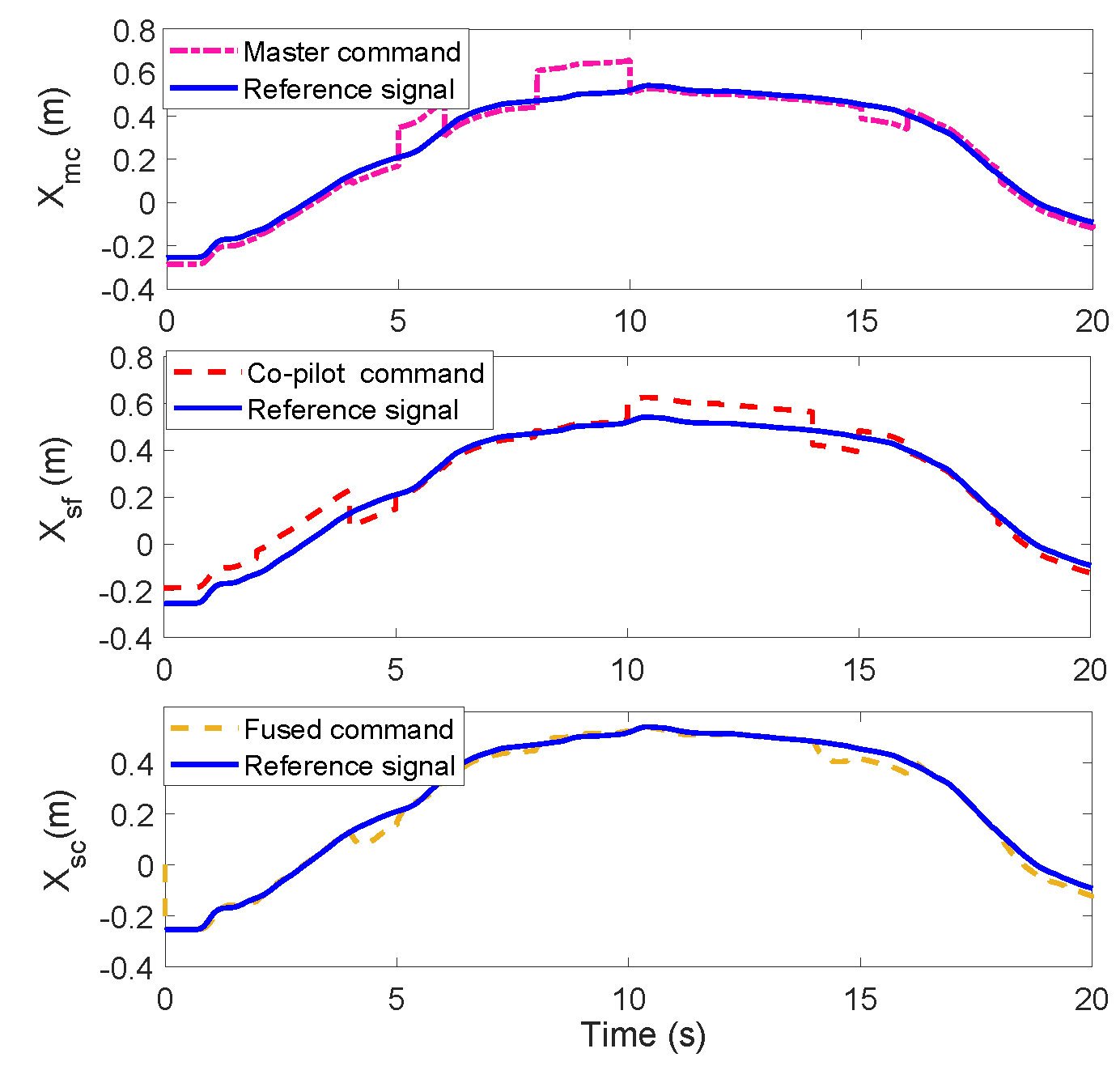}
\caption{The position tracking curves of the master operator and co-pilot.}
\vspace{-4mm}
\label{f:single-master}
\end{figure}
\subsection{Dual-Pilot Collaboration}
One master operator drew a circle trajectory based on visual feedback while the co-pilot with similar expertise carried out the same task. Two operators could communicate via local area network (LAN), leading to a negligible time delay. The same NN structure is utilised in two DDPG-based agents, where the critic NN is composed of three hidden layers with five neurons each, and one other layer with two neurons, while the actor NN is composed of one hidden layer with three neurons. The discount factor was set as 0.99, minibatch size as 100, experience buffer length as $10^4$, noise variance as 0.01 and variance decay rate as $10^{-5}$. The RL agents are trained using the generated trajectories for 20 episodes and each episode consisting of 2000 steps, where each step size is 0.01s. The boundaries of $[-1,1]$ and $[0,1]$ are set for the outputs of DDPG1 and DDPG2, respectively. The root-mean-square error (RMSE) is chosen as the evaluation criteria for operational performance. 

Due to page limitations, we only show the $x$-direction position profiles of master operator ($X_{mc}$), co-pilot ($X_{sf}$) and final command ($X_{sc}$) blended in the arbitration module, as shown in Fig.\,\ref{f:single-master}. Three subjective errors occurred during operation, resulting in the deviation of the trajectory (RMSE = 0.057569). The co-pilot was aware of the error at the master side during the corresponding three periods and compensated by improving its own trajectory. The co-pilot also made mistakes at other times, for example in the first five seconds and 10-15 seconds (RMSE = 0.058192). The final command $X_{sc}$ sent to the robot was generated through blending the master and co-pilot command in the DDPG-based arbitration (RMSE = 0.022571). It shows human subjective errors can be effectively overcome and the necessity of introducing co-pilot.

\subsection{Triple-Pilot Collaboration with Time Delay}
\begin{figure}[h]
\centering
\vspace{-4mm}
\includegraphics[width=8.5cm]{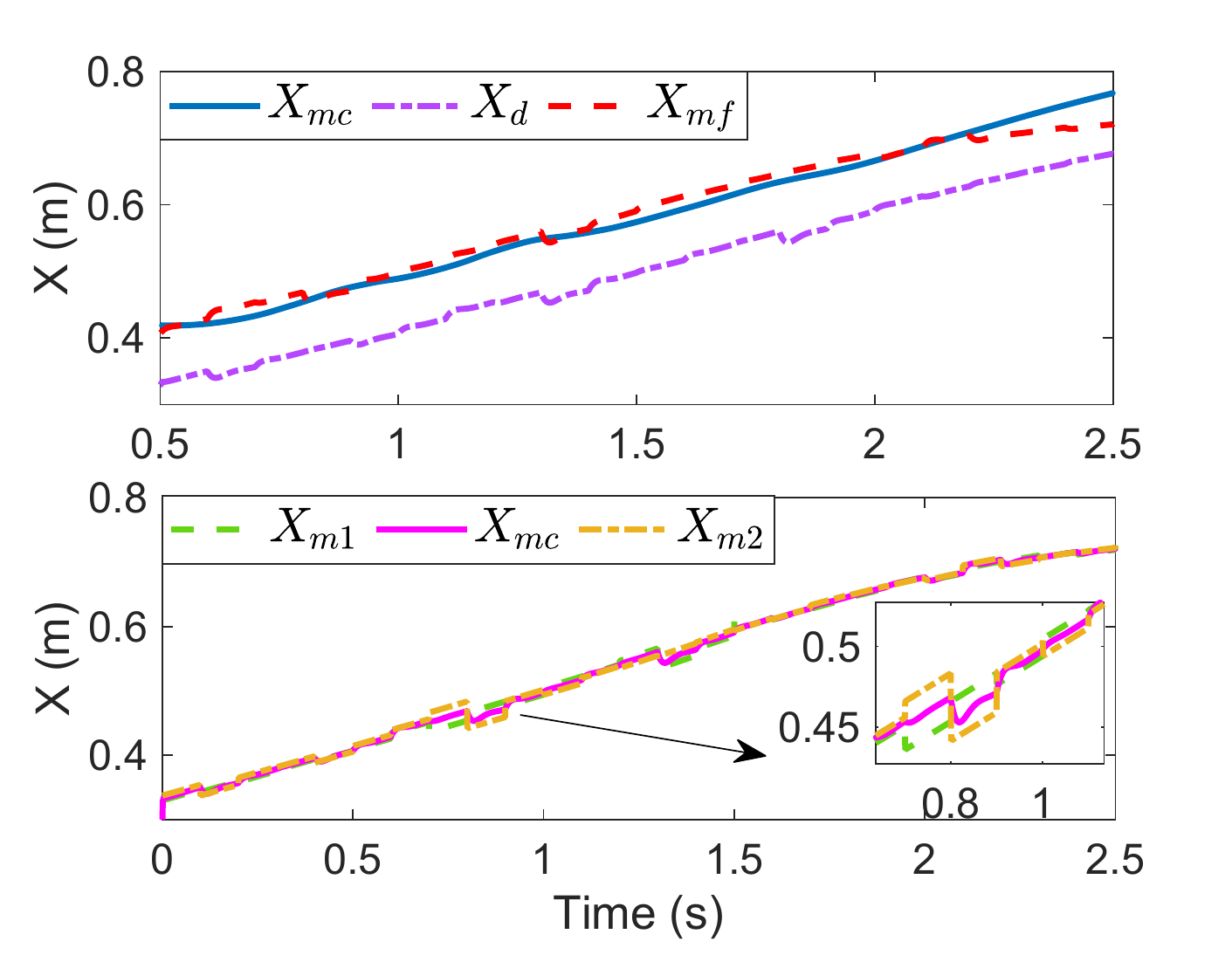}
\caption{The position commands for master and slave restore.}
\vspace{-2mm}
\label{f:dual-master}
\end{figure}

\begin{figure}
\centering
\includegraphics[width=8.5cm]{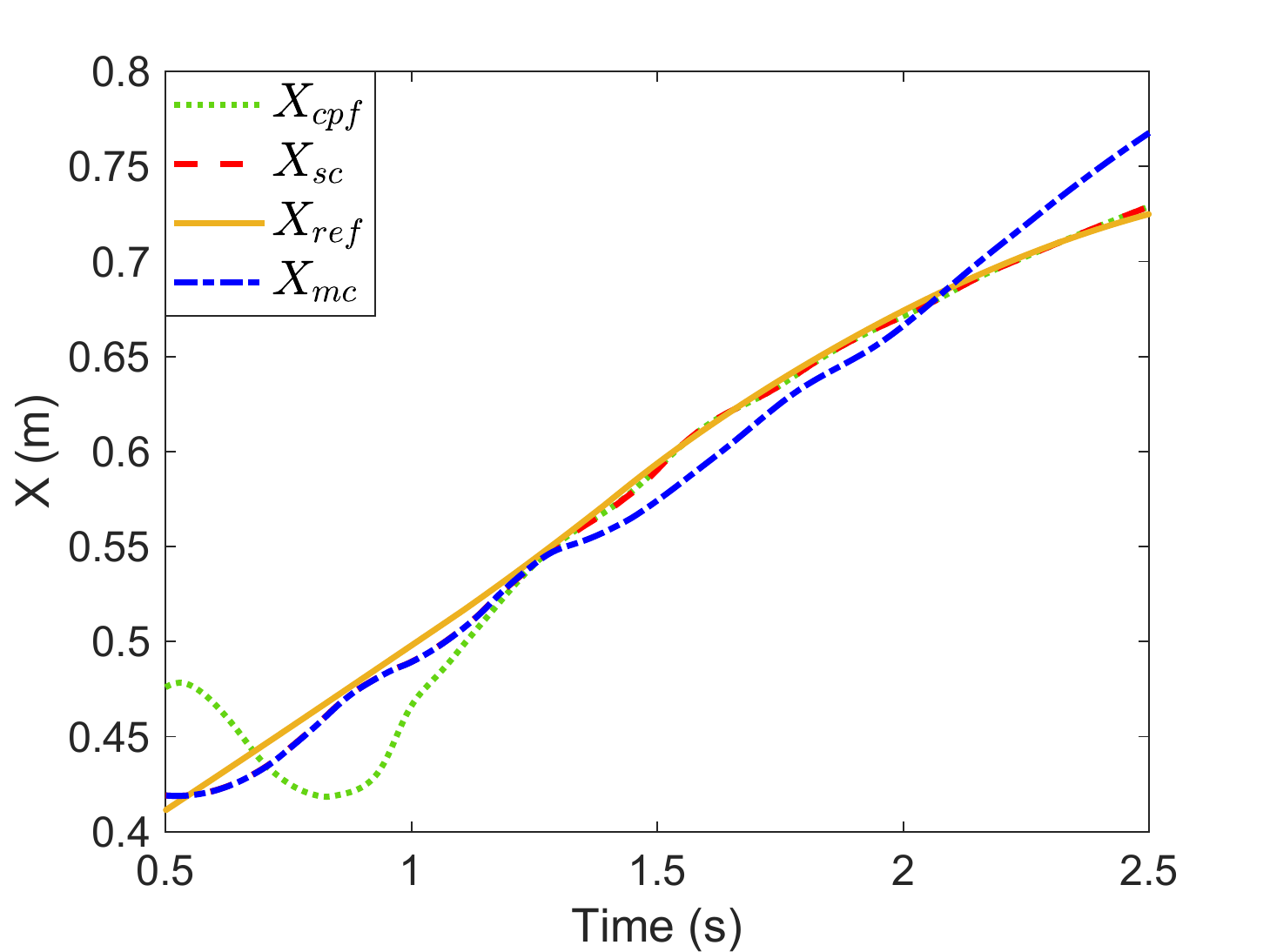}
\caption{Position commands of master operators and co-pilot.}
\vspace{-3mm}
\label{f:arbitration}
\end{figure}

\begin{figure}
\centering
\includegraphics[width=8.5cm]{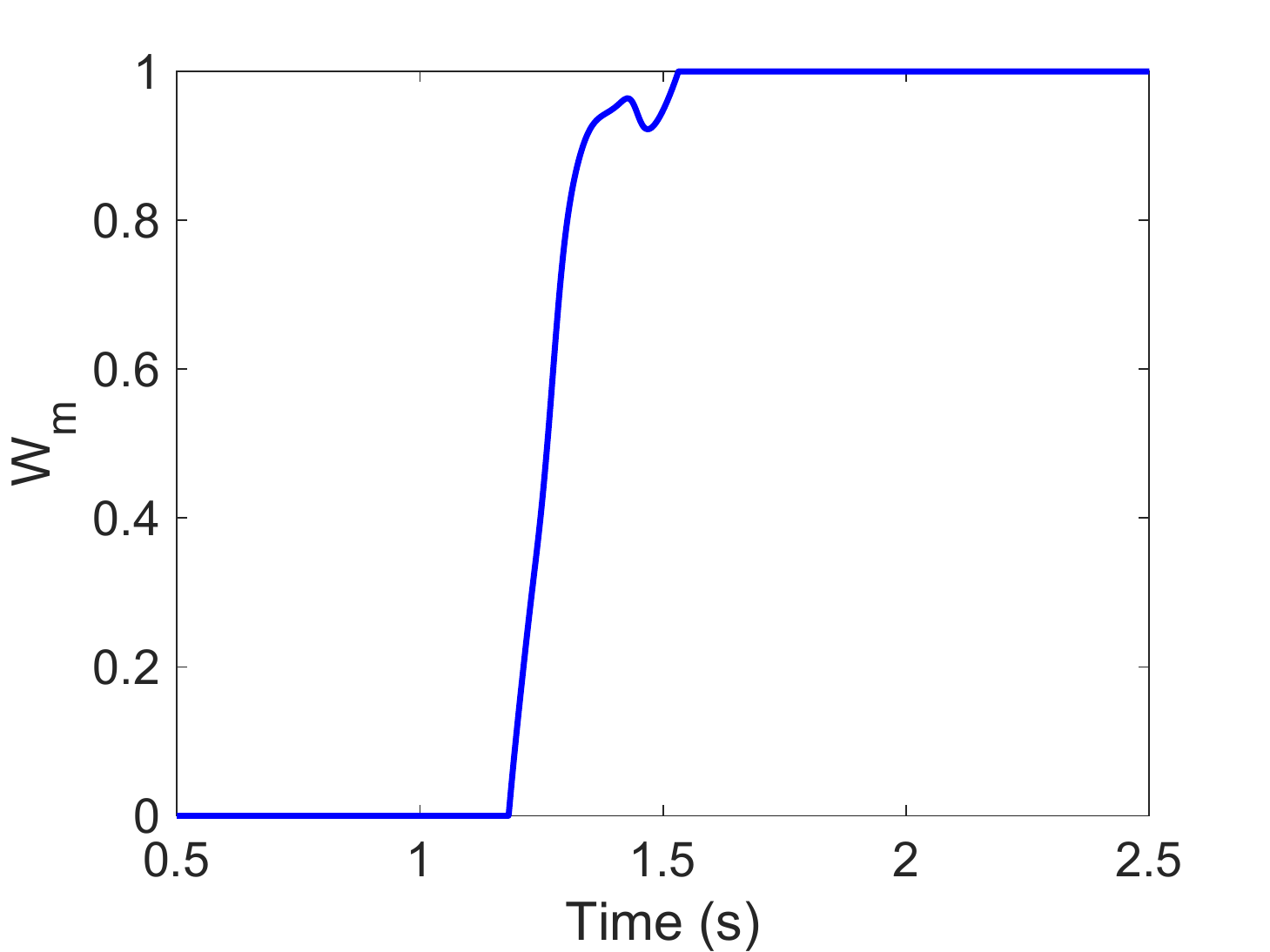}
\caption{Curve of arbitration adjustment.}
\vspace{-3mm}
\label{f:Wm}
\end{figure}

In this scenario, two operators performed a circle drawing operation at the master side with visual feedback and reconstructed force feedback. A co-pilot with similar experience carried out the same task at the slave side, where the master and slave communicated via TCP/IP protocol with a 0.5 second delay. The system control frequency is 100Hz.
The parameters of KF were set as follows:
$A=\Big[\begin{smallmatrix}
1 & 10^{-3} & 10^{-6}/2\\
0 & 1 & 10^{-3}\\
0 & 0 & 1
\end{smallmatrix}\Big]$, 
$W=\Bigg[\begin{smallmatrix}
10^{-12}/36 & 10^{-10}/12 & 10^{-8}/6\\
10^{-10}/12 & 10^{-8}/4 & 10^{-6}/2\\
10^{-8}/6 & 10^{-6}/2 & 10^{-4}/2
\end{smallmatrix}\Bigg]$, 
$P_0=\Big[\begin{smallmatrix}
10^{15} & 0 & 0\\
0 & 10^{15} & 0\\
0 & 0 & 10^{15}
\end{smallmatrix}\Big]$,
$C=\begin{bmatrix}
1 & 0 & 0
\end{bmatrix}$, $R=10^{-6}$. The rest parameters remain consistent with those in the above experiment. Due to the page limitation, we only show the $x$-direction data in the remaining figures. The position profiles of master and slave restore commands are depicted in Fig.\,\ref{f:dual-master}, which clearly shows that the DDPG1 effectively compensates for phase delays due to communication channel. The reliability of the master command is improved by combining the two operators' intents with the help of KF (RMSE between $X_{mc}$ and $X_{ref}$ is 0.04747). $X_{ref}$ stands for the task-dependent reference signal. Fig.\,\ref{f:arbitration} illustrates the decision making process of the master operators and co-pilot. At the beginning of the collaboration with dual master operators, the co-pilot deviated from the desired trajectory due to operational errors, therefore the weight assigned to him was zero, as shown in Fig.\,\ref{f:Wm}. In the middle of collaboration, the co-pilot compensated for subjective errors at the master side so that the corresponding weight was increased. The co-pilot and arbitration module allow dynamic changes in work efficiency at the master side, which guarantees the robustness of commands in the event of operational errors (RMSE between $X_{sc}$ and $X_{ref}$ is 0.0051387).

\begin{figure}
\centering
\includegraphics[width=8.5cm]{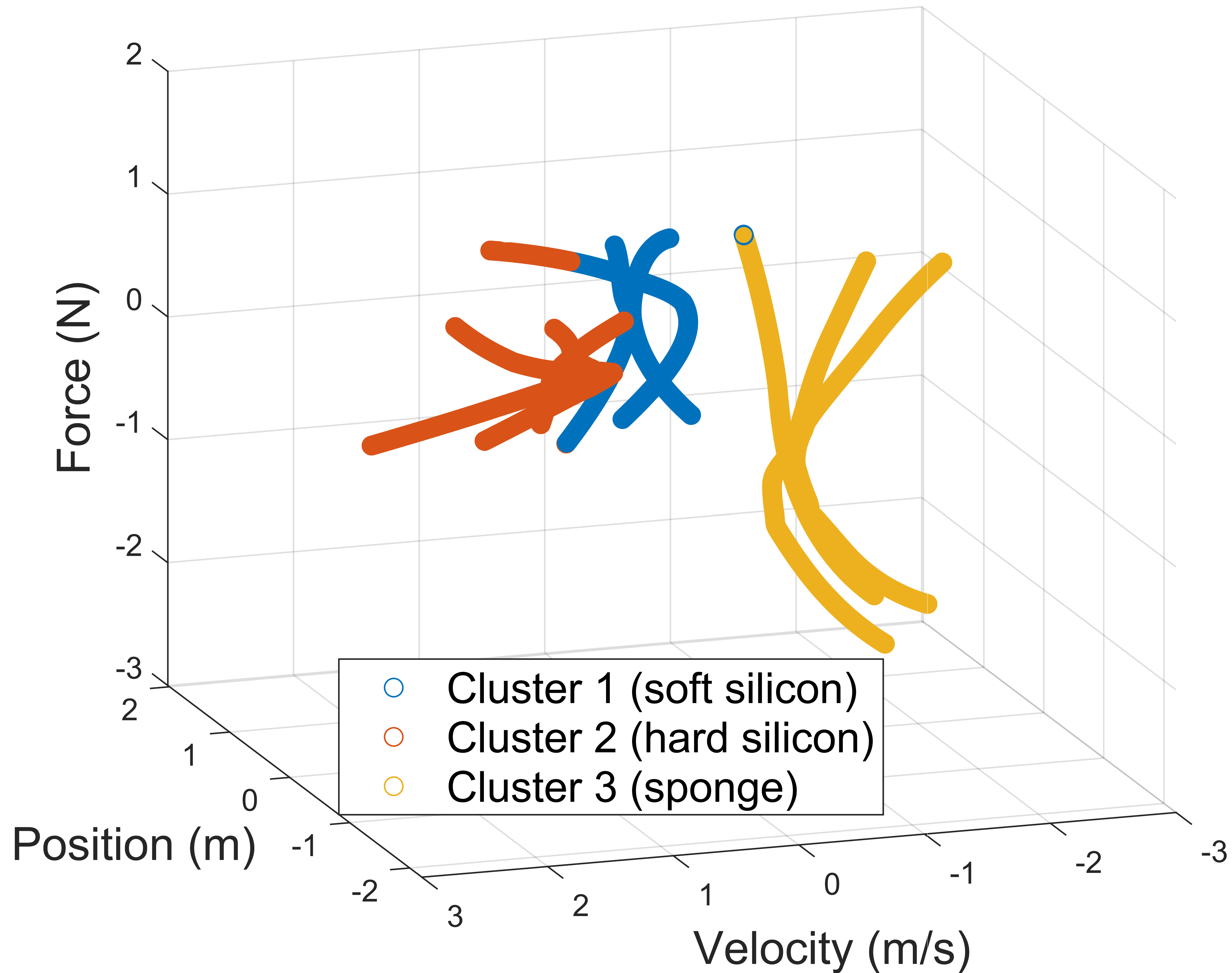}
\caption{Fuzzy clustering results.}
\vspace{-3mm}
\label{f:clustering_result}
\end{figure}

\begin{figure}
\centering
\includegraphics[width=8.2cm]{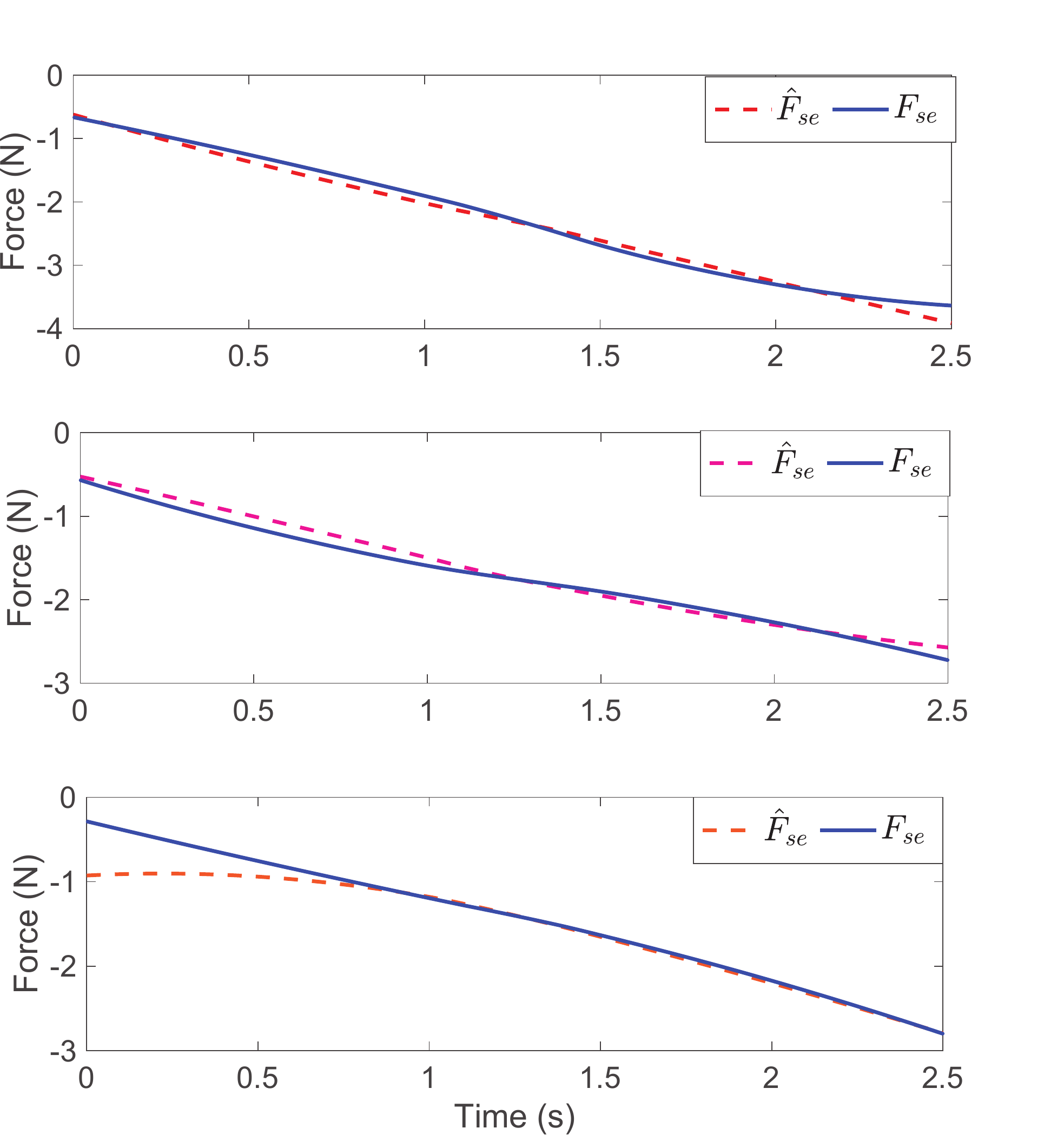}
\caption{Force estimation results.}
\vspace{-3mm}
\label{f:force-estimation}
\end{figure}

To validate the effectiveness of the proposed force estimator, we tested the performance of force estimation in three interaction environments (hard silicon, soft silicon and sponge) provided force feedback for local reconstruction to the master operators. We configured an ATI 6-axis force/torque sensor for the end-effector of the slave robot during the training phase. Fifty interaction experiments were conducted in each of the three material environments, where three trials were randomly selected as the training set and the rest as the test set. Fig.\,\ref{f:clustering_result} illustrates the clustering results used in the test. The average RMSEs for the interaction forces are 0.097904 (hard silicon), 0.088005 (soft silicon) and 0.18961 (sponge) respectively. The 
curves from top to bottom in Fig.\,\ref{f:force-estimation} show the estimated interaction forces versus the measured forces in hard silicon, soft silicon and sponge materials respectively. This shows that the designed force estimator effectively establishes a mapping between the input state and the force, i.e. it provides force feedback to the master in force-sensor-free scenarios. The operators can feel the force interaction between the end-effector and the environment without a sense of delay. The stiffness of the three materials can be also felt to be different according to the force feedback applied.

\section{Conclusion}
In this paper, we propose a new control framework for multi-pilot collaborative teleopertion. DDPG-based RL agents are employed to enable the reduction of the master intents and multi-source intent decisions. In order to enhance the master perception of the interaction environment, we develop a new interactive force estimator based on IT2 T-S fuzzy identification, which facilities the force feedback reconstructed at the master side. Two key metrics, tracking accuracy and telepresence, are thus guaranteed.
Experiments in scenarios of \textit{dual-pilot collaboration }and \textit{triple-pilot collaboration with time delay }verify the merits of the control.  It should be noted that the framework proposed is based on the premise that the task target is known, which is widely used in systems for training operators. Scenarios with unknown or time-varying task targets will be investigated in the future work.

\section*{Acknowledgement}
The authors appreciate Dr. Xiaoxiao Cheng, Pakorn Uttayopas and Sarah Chams Bacha from Imperial College for collecting the data of the experiments.





\bibliographystyle{IEEEtran}
\bibliography{IEEEabrv,Bibliography}

\begin{thebibliography}{10}
\providecommand{\url}[1]{#1}
\csname url@samestyle\endcsname
\providecommand{\newblock}{\relax}
\providecommand{\bibinfo}[2]{#2}
\providecommand{\BIBentrySTDinterwordspacing}{\spaceskip=0pt\relax}
\providecommand{\BIBentryALTinterwordstretchfactor}{4}
\providecommand{\BIBentryALTinterwordspacing}{\spaceskip=\fontdimen2\font plus
\BIBentryALTinterwordstretchfactor\fontdimen3\font minus
  \fontdimen4\font\relax}
\providecommand{\BIBforeignlanguage}[2]{{%
\expandafter\ifx\csname l@#1\endcsname\relax
\typeout{** WARNING: IEEEtran.bst: No hyphenation pattern has been}%
\typeout{** loaded for the language `#1'. Using the pattern for}%
\typeout{** the default language instead.}%
\else
\language=\csname l@#1\endcsname
\fi
#2}}
\providecommand{\BIBdecl}{\relax}
\BIBdecl

\bibitem{niemeyer1991stable}
G.~Niemeyer and J.-J. Slotine, ``Stable adaptive teleoperation,'' \emph{IEEE
  Journal of Oceanic Engineering}, vol.~16, no.~1, pp. 152--162, 1991.

\bibitem{lawrence1993stability}
D.~A. Lawrence, ``Stability and transparency in bilateral teleoperation,''
  \emph{IEEE Transactions on Robotics and Automation}, vol.~9, no.~5, pp.
  624--637, 1993.

\bibitem{wbai2021dual}
W.~Bai, N.~Zhang, B.~Huang, Z.~Wang, F.~Cursi, Y.-Y. Tsai, B.~Xiao, and E.~M.
  Yeatman, ``Dual-arm coordinated manipulation for object twisting with human
  intelligence,'' in \emph{2021 IEEE International Conference on Systems, Man,
  and Cybernetics (SMC)}.\hskip 1em plus 0.5em minus 0.4em\relax IEEE, 2021.

\bibitem{troccaz2019frontiers}
J.~Troccaz, G.~Dagnino, and G.-Z. Yang, ``Frontiers of medical robotics: from
  concept to systems to clinical translation,'' \emph{Annual Review of
  Biomedical Engineering}, vol.~21, pp. 193--218, 2019.

\bibitem{Wang2019Adaptive}
Z.~Wang, Z.~Chen, Y.~Zhang, X.~Yu, X.~Wang, and B.~Liang, ``Adaptive
  finite-time control for bilateral teleoperation systems with jittering time
  delays,'' \emph{International Journal of Robust and Nonlinear Control},
  vol.~29, no.~4, pp. 1007--1030, 2019.

\bibitem{arcara2002control}
P.~Arcara and C.~Melchiorri, ``Control schemes for teleoperation with time
  delay: A comparative study,'' \emph{Robotics and Autonomous systems},
  vol.~38, no.~1, pp. 49--64, 2002.

\bibitem{rezazadeh2019robotic}
S.~Rezazadeh, W.~Bai, M.~Sun, S.~Chen, Y.~Lin, and Q.~Cao, ``Robotic spinal
  surgery system with force feedback for teleoperated drilling,'' \emph{The
  Journal of Engineering}, vol. 2019, no.~14, pp. 500--505, 2019.

\bibitem{bai2017modular}
W.~Bai, Q.~Cao, P.~Wang, P.~Chen, C.~Leng, and T.~Pan, ``Modular design of a
  teleoperated robotic control system for laparoscopic minimally invasive
  surgery based on ros and rt-middleware,'' \emph{Industrial Robot: An
  International Journal}, 2017.

\bibitem{bai2017novel}
W.~Bai, Q.~Cao, C.~Leng, Y.~Cao, M.~G. Fujie, and T.~Pan, ``A novel optimal
  coordinated control strategy for the updated robot system for single port
  surgery,'' \emph{The International Journal of Medical Robotics and Computer
  Assisted Surgery}, vol.~13, no.~3, p. e1844, 2017.

\bibitem{xue2021Continuous}
T.~Xue, Z.~Wang, T.~Zhang, O.~Bai, M.~Zhang, and B.~Han, ``Continuous
  finite-time torque control for flexible assistance exoskeleton with delay
  variation input,'' \emph{Robotica}, vol.~39, no.~2, p. 291–316, 2021.

\bibitem{Xue2019Adaptive}
T.~{Xue}, Z.~{Wang}, T.~{Zhang}, and M.~{Zhang}, ``Adaptive oscillator-based
  robust control for flexible hip assistive exoskeleton,'' \emph{IEEE Robotics
  and Automation Letters}, vol.~4, no.~4, pp. 3318--3323, 2019.

\bibitem{Ivanova2021Short}
E.~Ivanova, J.~Eden, S.~Zhu, G.~Carboni, A.~Yurkewich, and E.~Burdet, ``Short
  time delay does not hinder haptic communication benefits,'' \emph{IEEE
  Transactions on Haptics}, vol.~14, no.~2, pp. 322--327, 2021.

\bibitem{Wang2020Fault-Tolerant}
Z.~Wang, B.~Liang, Y.~Sun, and T.~Zhang, ``Adaptive fault-tolerant
  prescribed-time control for teleoperation systems with position error
  constraints,'' \emph{IEEE Transactions on Industrial Informatics}, vol.~16,
  no.~7, pp. 4889--4899, 2020.

\bibitem{Wang2020Event}
Z.~{Wang}, H.~{Lam}, B.~{Xiao}, Z.~{Chen}, B.~{Liang}, and T.~{Zhang},
  ``Event-triggered prescribed-time fuzzy control for space teleoperation
  systems subject to multiple constraints and uncertainties,'' \emph{IEEE
  Transactions on Fuzzy Systems}, vol.~29, no.~9, pp. 2785--2797, 2021.

\bibitem{SUN2014Application}
D.~Sun, F.~Naghdy, and H.~Du, ``Application of wave-variable control to
  bilateral teleoperation systems: A survey,'' \emph{Annual Reviews in
  Control}, vol.~38, no.~1, pp. 12--31, 2014.

\bibitem{nuno2011passivity}
E.~Nu{\~n}o, L.~Basa{\~n}ez, and R.~Ortega, ``Passivity-based control for
  bilateral teleoperation: A tutorial,'' \emph{Automatica}, vol.~47, no.~3, pp.
  485--495, 2011.

\bibitem{uddin2016predictive}
R.~Uddin and J.~Ryu, ``Predictive control approaches for bilateral
  teleoperation,'' \emph{Annual Reviews in Control}, vol.~42, pp. 82--99, 2016.

\bibitem{Chan2014Application}
L.~Chan, F.~Naghdy, and D.~Stirling, ``Application of adaptive controllers in
  teleoperation systems: A survey,'' \emph{IEEE Transactions on Human-Machine
  Systems}, vol.~44, no.~3, pp. 337--352, 2014.

\bibitem{shahbazi2018systematic}
M.~Shahbazi, S.~F. Atashzar, and R.~V. Patel, ``A systematic review of
  multilateral teleoperation systems,'' \emph{IEEE transactions on haptics},
  vol.~11, no.~3, pp. 338--356, 2018.

\bibitem{Khademian2012Dual-User}
B.~{Khademian} and K.~{Hashtrudi-Zaad}, ``Dual-user teleoperation systems: New
  multilateral shared control architecture and kinesthetic performance
  measures,'' \emph{IEEE/ASME Transactions on Mechatronics}, vol.~17, no.~5,
  pp. 895--906, 2012.

\bibitem{lillicrap2015continuous}
T.~P. Lillicrap, J.~J. Hunt, A.~Pritzel, N.~Heess, T.~Erez, Y.~Tassa,
  D.~Silver, and D.~Wierstra, ``Continuous control with deep reinforcement
  learning,'' \emph{arXiv preprint arXiv:1509.02971}, 2015.

\bibitem{chiu1994fuzzy}
S.~L. Chiu, ``Fuzzy model identification based on cluster estimation,''
  \emph{Journal of Intelligent \& Fuzzy Systems}, vol.~2, no.~3, pp. 267--278,
  1994.

\bibitem{Mendel2006it2}
J.~M. Mendel, R.~I. John, and F.~Liu, ``Interval type-2 fuzzy logic systems
  made simple,'' \emph{IEEE Trans. Fuzzy Syst.}, vol.~14, no.~6, pp. 808--821,
  2006.

\end{thebibliography}
%


\end{document}